\documentclass[letterpaper, 10 pt, conference]{ieeeconf}  
\IEEEoverridecommandlockouts                              
\usepackage{graphics} 
\usepackage{epsfig} 
\usepackage{mathptmx} 
\usepackage{times} 

\usepackage{amssymb} 
\usepackage{mathtools}
\usepackage{graphics} 
\usepackage{epsfig} 
\usepackage{amsmath, amssymb} 
\usepackage{booktabs} 
\usepackage{hyperref}
\usepackage{multirow} 
\usepackage{comment}

\title{\LARGE \bf
3D Annotation Of Arbitrary Objects In The Wild
}

\author{Kenneth Blomqvist$^{1}$ and Julius Hietala$^{1}$ 
\thanks{Both authors are with Stray Robots and contributed equally %
{\tt\small ken@strayrobots.io} and {\tt\small julius@strayrobots.io}}%
}

\begin{document}

\maketitle
\thispagestyle{empty}
\pagestyle{empty}

\begin{abstract}
Recent years have produced a variety of learning based methods in the context of computer vision and robotics. Most of the recently proposed methods are based on deep learning, which require very large amounts of data compared to traditional methods. The performance of the deep learning methods are largely dependent on the data distribution they were trained on, and it is important to use data from the robot's actual operating domain during training. Therefore, it is not possible to rely on pre-built, generic datasets when deploying robots in real environments, creating a need for efficient data collection and annotation in the specific operating conditions the robots will operate in. The challenge is then: how do we reduce the cost of obtaining such datasets to a point where we can easily deploy our robots in new conditions, environments and to support new sensors? As an answer to this question, we propose a data annotation pipeline based on SLAM, 3D reconstruction, and 3D-to-2D geometry. The pipeline allows creating 3D and 2D bounding boxes, along with per-pixel annotations of arbitrary objects without needing accurate 3D models of the objects prior to data collection and annotation. Our results showcase almost 90\% Intersection-over-Union (IoU) agreement on both semantic segmentation and 2D bounding box detection across a variety of objects and scenes, while speeding up the annotation process by several orders of magnitude compared to traditional manual annotation. 

\end{abstract}

\section{INTRODUCTION}

The majority of computer vision methods used in robotics today are supervised learning based and require vasts amounts of labeled training examples to fit parameters \cite{sunderhauf2018limits}. Many approaches have been devised to learn in an unsupervised or self-supervised fashion to avoid having to annotate data. However, any such methods can still benefit from annotated examples. 

The majority of computer vision methods process one image frame at the time. They typically find features or make predictions from single images. Learning based methods would typically predict higher level features such as keypoints, heatmaps, bounding boxes or semantic segmentation maps. The annotated examples they consume are usually labeled one at the time using a tool such as LabelMe \cite{russell2008labelme} where single images are annotated one at the time with the prediction targets. A large industry has even developed around such tools for companies to outsource the labeling effort to large teams of workers.

While labeling single images one at the time might be feasible for tasks and robots that have massive markets, such as autonomous driving, the long tail of robotic applications can't afford to invest in annotating the millions of examples required to train modern deep learning methods to perfection. This is made worse by the fact that previously labeled data might become useless if a sensor is changed or the robot is deployed in a different environment. 

3D computer vision methods \cite{ioannidou2017deep} that make predictions directly in 3D space have been developed. For these applications, relying on single 2D labeled image examples is not possible and the depth dimension also needs to be annotated. Humans have a deeply ingrained spatial understanding of the world from a lifetime of experience observing the world. Computer vision algorithms usually mostly start from scratch. Baking this type of knowledge into algorithms could be a path to more efficient and precise systems and having 3D annotated scenes could greatly facilitate such research directions.

\begin{figure}[t]
    \centering
    \includegraphics[width=\linewidth]{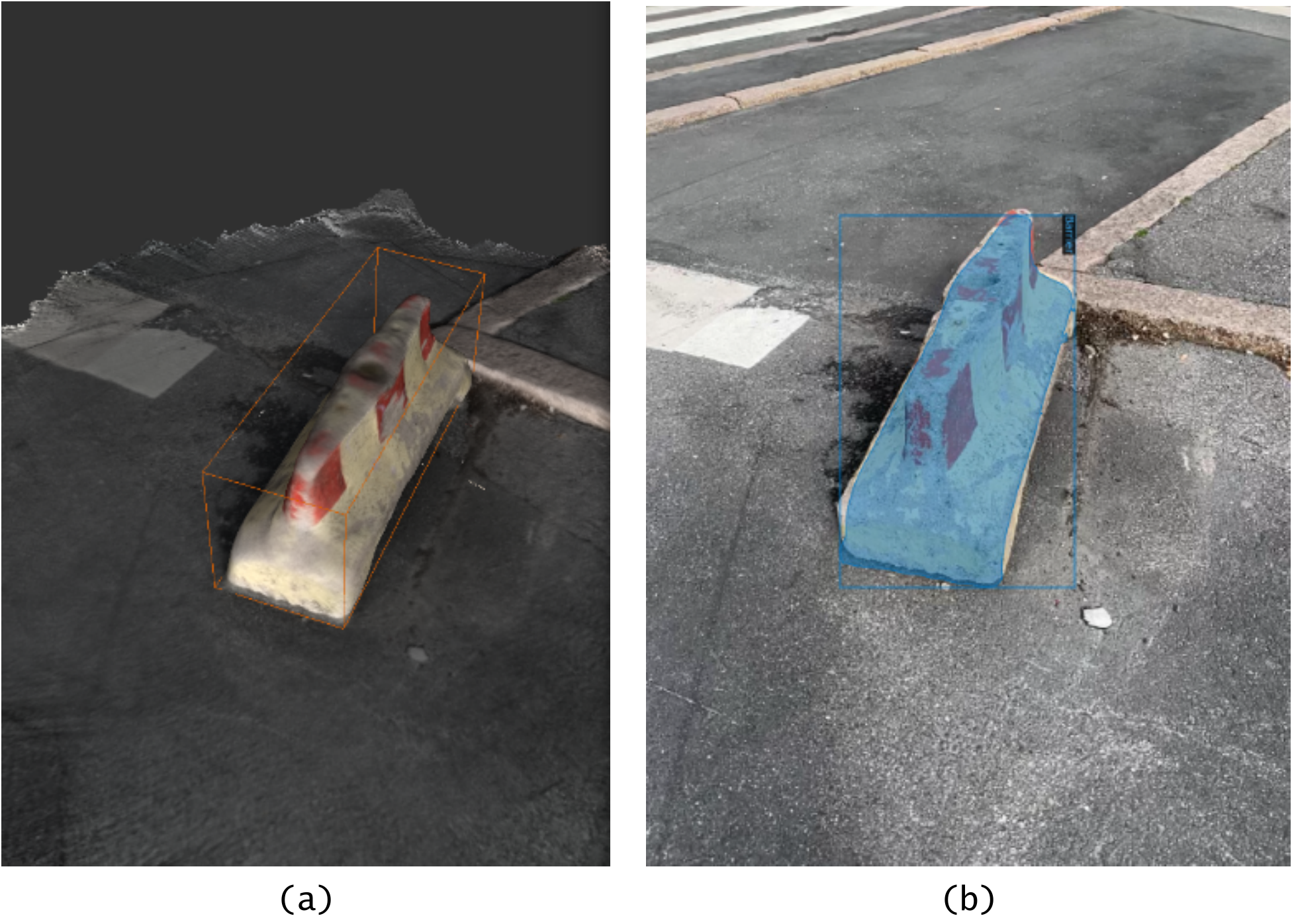}
    \caption{(a) Objects are annotated in a 3d graphical user interface with bounding boxes. (b) Bounding boxes and per-pixel labels are projected back to the original images. }
    \label{fig:studio}
\end{figure}

LabelFusion \cite{marion2018label} presented a 3D data annotation system based on dense environment maps for RGB-D data. It can be used to compute 6D pose labels for objects with a known model. In most cases, object models are not available, intra-category variation or the fact that objects deform, can make relying on accurate object models impossible. Having a tool that could very quickly provide ground-truth labels in such cases, could allow us to deploy more powerful learning-based algorithms, where it previously hasn't been possible.

In this paper, we present a general toolkit\footnote{We have made the tool available for download and use at \href{https://docs.strayrobots.io/installation/index.html}{https://docs.strayrobots.io}} for annotating 3D scenes with semantic information. The proposed toolkit takes as input a stream of RGB-D images, reconstructs the scene into a global 3D dense map and recovers camera poses for each image. Users of the tool can annotate the scene with object and other semantic information. Based on the scene reconstruction, camera poses and annotations, our tool is able to generate datasets for 2D or 3D object detection, semantic segmentation and keypoint detection tasks. 

We demonstrate the tool's effectiveness on a set of experiments in real-world scenes and compare our method to a dataset that has been labeled in 2D one image at the time, both in accuracy and the time it takes to create the labels. Additionally, we present qualitative results of the produced labels and evaluate detection model performance that has been trained on the examples to validate the adequacy of our method.

To summarize our contributions are as follows:
\begin{itemize}
    \item A semi-automatic 3D RGB-D data annotation tool that speeds up annotation by several orders of magnitude, while maintaining an acceptable level of label accuracy
    \item Quantitative comparison of the generated labels against manually annotated examples and performance evaluation in a real-world detection task
    \item Qualitative results that showcase the accuracy of the generated labels as well as models trained on those labels
\end{itemize}

\section{RELATED WORK}
\begin{figure*}[t!]
    \includegraphics[width=\linewidth]{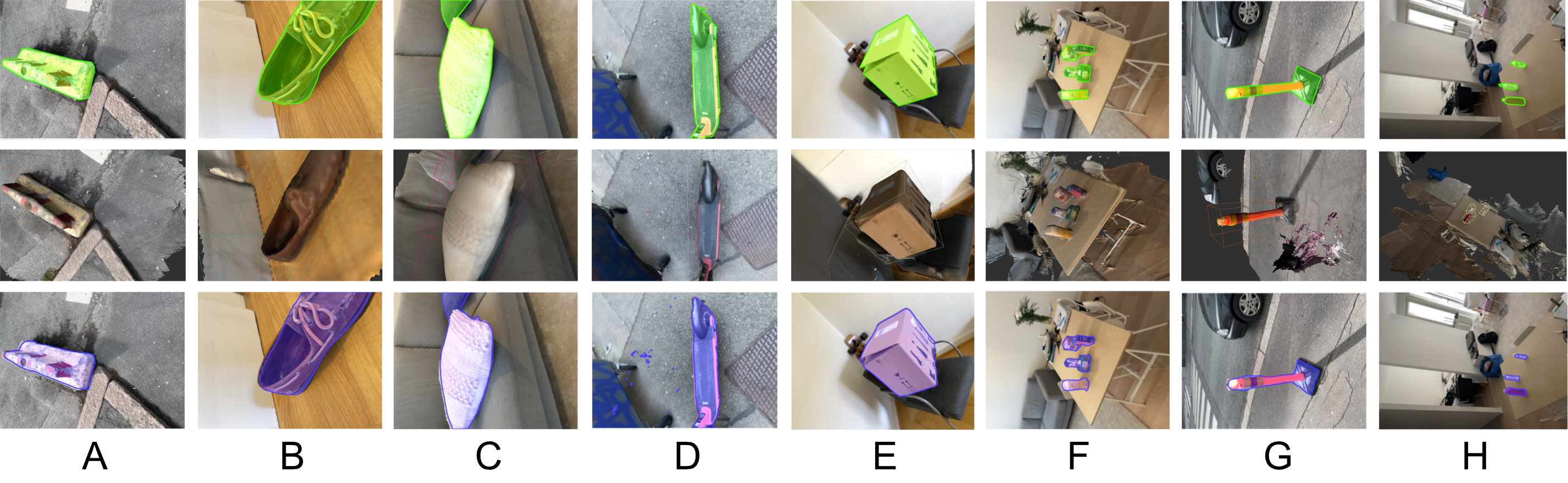}
    \centering
    \caption{All classes in our dataset visualized: Barriers (A), Shoes (B), Pillows (C), Scooters (D), Boxes (E), Food items (F), Poles (G), Bottles (H). The first row in green shows the hand labeled human semantic segmentation mask. The second row shows the scene reconstruction in the graphical user interface. The third row in blue shows segmentation masks, generated by our method.}
    \label{fig:masks}
\end{figure*}

\subsection{2D Annotation and Active Learning}

For computer vision tasks such as object detection, keypoint detection or semantic segmentation a tool such as LabelMe \cite{russell2008labelme} or similar is usually used to annotate images one-by-one by directly drawing on the image. 

Methods such as Deep Extreme Cut \cite{maninis2018deep}, have been developed to reduce the amount of manual work required to create the segmentation masks. It reduces the problem of creating semantic segmentation masks to a keypoint annotation problem, and generates segmentation masks from a few extreme points. 

Another way to speed up data annotation is a active learning \cite{settles2011theories}. ViewAL \cite{siddiqui2020viewal} takes this type of approach and uses viewpoint entropy as an active learning acquisition function to achieve almost undisturbed performance using only a fraction of the training data. Such methods could very well be combined with our 3D labeling approach to make use of the different viewpoints we recover in the preprocessing step, to further reduce the labeling burden.

Another way to speed up labeling was explored by \cite{hansen2021novel} where they fit a model while the user is labeling images to modify the task from an active annotation task to one of checking and adjusting proposals. 

\subsection{3D Annotation}

Some tasks are inherently 3D in nature, which has led to the development of tools that not only annotate single images in 2D, but can also annotate the depth dimension.

ScanNet \cite{dai2017scannet} presented a dataset and tool for annotating dense 3D scenes with semantic information. The tool similarly creates a surface reconstruction, but then requires users to annotate objects by aligning CAD models of the objects. 

LabelFusion \cite{marion2018label} pioneered 3D data annotation for pose estimation by providing a tool for generating ground truth object pose datasets with semantic segmentation maps, by leveraging a dense mapping system. However, LabelFusion does require using high definition 3D object models. 3D object models are often not available and obtaining such models requires using an expensive object scanning setup. Furthermore, objects might be too large to scan, they might have category-level variation requiring a mesh for each variant of the object, objects might be articulated or deformable, making an object model based approach harder to apply. Our method operates without object models.

EasyLabel \cite{suchi2019easylabel} introduced a semi-automatic method for obtaining instance segmentations of 3D scenes. The method builds on the idea of incrementally building up the scene by placing objects into the scene one at the time. While generating very accurate labels, such methods can not be applied in the general case where a robot is moving through an uncontrolled environment.

SAnE \cite{arief2020sane} presents a semi-automatic tool that makes annotating point clouds for autonomous driving more efficient. The core of their tool is a guided tracking algorithm that can propagate object labels over time and it can work on dynamic scenes. It only processes point cloud data and does not deal with the full problem of RGB-D annotation and it doesn't compute dense segmentation masks. A similar guided tracking algorithm could be used to extend our approach to dynamic scenes with moving objects.

SALT \cite{stumpf2021salt} proposes to use a GrabCut based approach to speed up labeling of RGB-D data. Similarly, their method does not require object or environment models. Their method operates directly on point clouds and does not build up a dense map of the scene.

Objectron \cite{ahmadyan2021objectron} recently introduced a sizeable dataset for 3D object detection. The dataset was collected on smartphones and similarly uses short video clips and leverages a SLAM pipeline for tracking the pose of the camera. However, Objectron does not use or include depth maps, nor does their collection method compute object segmentation masks, they don't provide a mesh of the scenes nor do they make their data acquisition tool available to the public.

\section{METHOD}

Our objective is to obtain ground truth labels for each frame in a set of ordered RGB-D frames that have been recorded by a conventional depth camera. The ground truth labels can be semantic information, such as semantic segmentation maps, 2D object detection bounding boxes or semantic keypoints in the scene. We propose to solve this problem by recovering the 3D structure of the scene and relying on a human annotator and graphical user interface to provide the semantic information within this 3D representation.

As a preprocessing step, we compute:
\begin{itemize}
    \item The trajectory of the camera with poses for each frame, relative to the first image
    \item A triangle mesh of the scene in the coordinate system of the first frame
\end{itemize}
These are used as input to the graphical user interface.

\subsection{Localization, Mapping and Dense Reconstruction}

As a first step, we extract camera poses for each color and corresponding depth image in a simultaneous localization and mapping (SLAM) step. To do this, we use the ORB-SLAM3 \cite{campos2021orb} RGB-D SLAM pipeline. We denote the transformation matrices $^{C_i}\mathbf{T}^W$, which transforms coordinates from world to the camera frame of the image at index $i$. The world frame is taken to be first image's camera frame. We denote quaternion orientations $\prescript{C_i}{}{\mathbf{q}}^W$ the rotation from world frame to frame $C_i$.

We take the trajectory computed by the SLAM step and integrate each color and depth image pair into a TSDF voxel grid \cite{curless1996volumetric, newcombe2011kinectfusion} and extract a mesh using the marching cubes algorithm \cite{lorensen1987marching}. For both TSDF integration and mesh extraction, we use the implementations available in Open3D \cite{zhou2018open3d}.

\subsection{GUI-based annotation}

Once the trajectory and scene geometry has been computed, we load them into our Stray Studio graphical user interface. The interface, visible in Figure \ref{fig:studio}, presents the 3D scene to the user. The user can pan and translate the camera to view the scene from different viewpoints. 

Users can add bounding boxes into the scene by clicking on the desired position. We ray trace the position on the mesh in the scene and place the box on the location that was clicked. The user can then adjust the position and resize it to encompass target objects in the scene.

\subsection{Label Extraction}

Once all objects in the scenes have been annotated with bounding boxes, our tool can extract labels for a variety of tasks, including 2D bounding box detection, 3D bounding box detection and semantic segmentation.

Here we denote the set of scene mesh vertices that are inside an object bounding box as $\mathbf{V}$ which is a set of 3D vectors. 

For 2D bounding box detection, for each frame we compute the top left point of the 2D object bounding box $\mathbf{l}_{i}$ and bottom right point of the bounding box $\mathbf{r}_{i}$ for frame $i$ using:
\begin{equation}
\begin{aligned}
    \mathbf{l}_{i} &= \min_{\mathbf{v}_j \in \mathbf{V}}{\mathbf{P}\mathbf{^{C_i}T^W} \mathbf{v}_j} \\
    \mathbf{r}_{i} &= \max_{\mathbf{v}_j \in \mathbf{V}}{\mathbf{P}\mathbf{^{C_i}T^W} \mathbf{v}_j}
\end{aligned}
\end{equation}
where $\mathbf{P}$ is the camera projection matrix, the minimum and maximum are taken element-wise. 

For 3D bounding box detection targets we transform each object bounding box into the camera frame by transforming the bounding box center $^{W}\mathbf{c}$ and orientation $\prescript{W}{}{\mathbf{o}}$ into each camera frame $C_i$
\begin{equation}
\begin{aligned}
    \prescript{C_i}{}{\mathbf{c}} &= ^{C_i}\mathbf{T}^W \prescript{W}{}{\mathbf{c}} \\
    \prescript{C_i}{}{\mathbf{o}} &= \prescript{C_i}{}{\mathbf{q}}^W \prescript{W}{}{\mathbf{o}}
\end{aligned}
\end{equation}
where $\mathbf{o}$ denotes the orientation of the bounding box. 

To obtain semantic segmentation masks, we cut out the vertices belonging to the object from the mesh ($\mathbf{V}$) with the corresponding faces and camera parameters and render masks for each frame using an OpenGL based pipeline.

\section{EXPERIMENTS}

We evaluate the proposed data annotation method on 34 different real-world scenes, with different backgrounds and lighting conditions and a varying amount of objects instances across 8 categories, totaling 56 616 images. Examples from each category are shown in Fig~\ref{fig:masks}. The objects in the categories vary in size and the amount of detail, and the scenes include shots taken from a range of distances to the target object. 

Data was collected using Apple iPhone 12 Pro smartphones, which are equipped with a time-of-flight depth sensor. Frames are captured at 60 frames per second. RGB frames have a resolution of 1920 by 1440 pixels and depth frames have a lower resolution of 256 by 192 pixels, which are upsampled using nearest neighbor interpolation to match the RGB frames.

\subsection{Label Accuracy}
 
\begin{table}[h]
\centering
\begin{tabular}{l | c c | c c} 
& \multicolumn{2}{c|}{segmentation} & \multicolumn{2}{|c}{bounding box}\\
 \toprule
 \bfseries category & \bfseries mean & \bfseries std & \bfseries mean & \bfseries \bfseries std \\ [0.5ex] 
 \toprule
 barrier & 0.90 & 0.029 & 0.90 & 0.057 \\
 \midrule
 shoe & 0.89 & 0.042 & 0.88 & 0.066\\
 \midrule
 box & 0.94 & 0.025 & 0.93 & 0.024\\
 \midrule
 scooter & 0.81 & 0.060 & 0.84 & 0.17\\
 \midrule
 bottle & 0.78 & 0.033 & 0.79 & 0.081\\
 \midrule
 food item & 0.82 & 0.14 & 0.85 & 0.15\\
 \midrule
 pillow & 0.91 & 0.034 & 0.94 & 0.038\\
 \midrule
 pole & 0.82 & 0.026 & 0.83 & 0.093\\
 \toprule
 \textbf{all} & 0.86 & 0.087 & 0.86 & 0.11\\
 \bottomrule
\end{tabular}
\caption{iou of labels from the proposed method compared to a hand labeled dataset}
\label{tab:iou}
\end{table}

\begin{figure}
    \includegraphics[width=\linewidth]{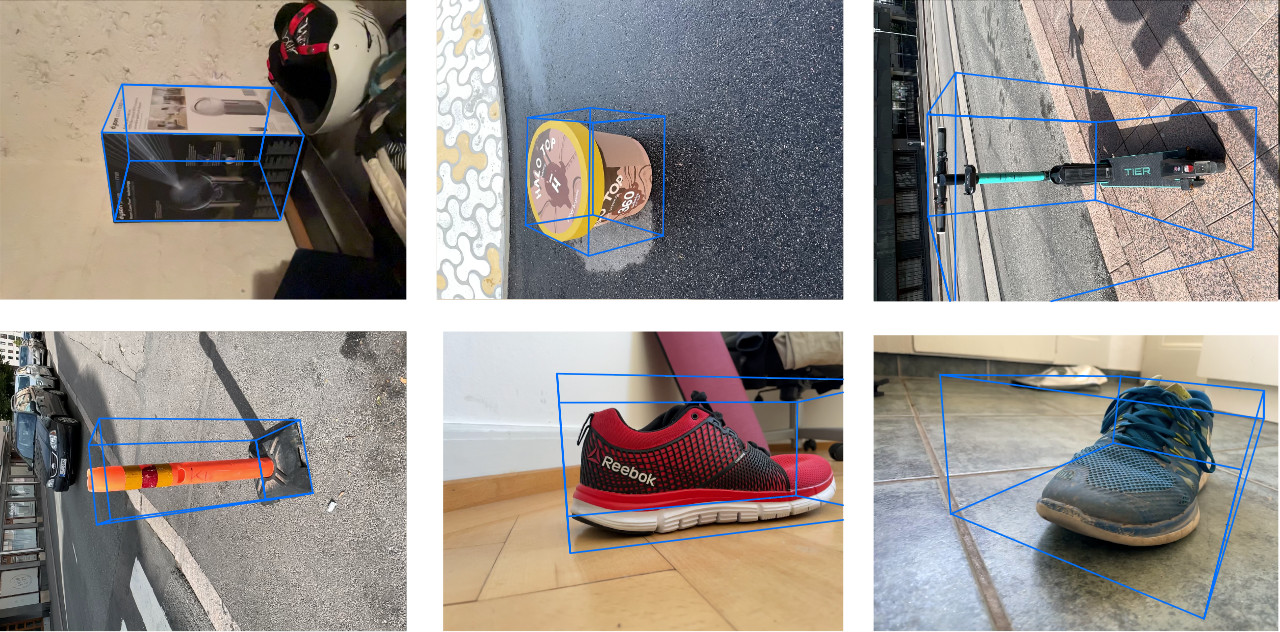}
    \caption{The annotated bounding boxes rendered on top of images sampled from our dataset.}
    \label{fig:boxes}
\end{figure}

After collecting the data, we run the reconstruction pipeline of the proposed method for each scene, and open each scene in the graphical user interface. We add a bounding box for each object of interest, after which all the label types can be projected back onto the 2D images. The bounding boxes in the user interface need only be approximate, as the projected labels are based on the the object mesh that is inside the bounding box. This not only achieves tighter bounding boxes, but also significantly speeds up the manual part of the annotation process, as the object dimensions do not need to be accurately annotated.

Our method has three main sources of error: localization error in the mapping stage of our pipeline, imperfectly reconstructed geometry and camera calibration error. 

To evaluate the quality of the per-pixel segmentation masks and 2D bounding boxes, we randomly sampled 458 images from our dataset and annotated the segmentation masks of the objects one image at a time using a traditional 2D image annotation tool \cite{labelme2016}, where segmentation masks are created by defining a polygon for each object in the frame. 2D bounding boxes are inferred from the minimum and maximum values of the resulting segmentation masks. The hand labeled examples serve as ground truth labels, against which we compare the 2D bounding boxes and per-pixel segmentation masks of the proposed method using the Intersection over Union (IoU) metric, also known as the Jaccard index. 

For the 3D pose annotations, it is not easy to obtain a similar ground truth dataset as in the 2D case. To get the ground truth 6DOF pose of the object, one would need to use an external motion capture system with markers at precisely known locations on the object and the camera or to measure the pose by hand relative to the camera for each frame, which is hard to do accurately and not possible in our moving handheld setting. Therefore we opt to evaluate them qualitatively.

We analyze the time required for label creation, and compare to the amount of time required by a human annotator using a 2D semantic segmentation tool.
\subsection{Downstream Task}

To see how well our labels do in practice on a downstream task, we evaluate the effect of using imperfect labels by training a state-of-the-art instance segmentation model with labels produced by our method. After training, we run inference on unseen images from unseen settings, and find out whether any errors in the annotations produced by the proposed method are visible in the predictions or whether they average out during model training.

\begin{table}[t]
\centering
\begin{tabular}{l | c  c  c  c  c  c} 
 & \multicolumn{2}{c}{Segmentation} & \multicolumn{2}{c}{Bounding Box}\\
 \toprule
 \bfseries Category & \bfseries mean & \bfseries std & \bfseries mean & \bfseries std \\ [0.5ex] 
 \toprule
 Barrier & 0.90 & 0.0075 & 0.95 & 0.022 \\
 \midrule
 Shoe & 0.86 & 0.11 & 0.82 & 0.12\\
 \midrule
 Box & 0.95 & 0.0040 & 0.92 & 0.014\\
 \midrule
 Scooter & 0.63 & 0.060 & 0.74 & 0.0075\\
 \midrule
 Bottle & 0.82 & 0.013 & 0.77 &  0.022\\
 \midrule
 Food item & 0.74 & 0.11 & 0.79 &  0.089\\
 \midrule
 Pillow & 0.90 & 0.031 & 0.80 & 0.029\\
 \midrule
 Pole & 0.76 & 0.048 & 0.77 & 0.056\\
 \toprule
 \textbf{All} & 0.81 & 0.13 & 0.81 & 0.13\\
 \bottomrule
\end{tabular}
\caption{IoU of segmentation model predictions against the ground truth labels.}
\label{tab:iou-train}
\end{table}

\section{RESULTS}

\subsection{Label quality}

Table~\ref{tab:iou} shows the mean and standard deviation of the IoU metric compared against hand labeled instances, for each category and annotation type. The mean of the per-pixel segmentation IoU for all categories is 86\%, which is slightly higher than what is reported by \cite{marion2018label}, despite the fact that our method does not require object models. However, the evaluation might not be fair, since \cite{marion2018label} only reports the metric on two random samples. The mean IoU for 2D bounding boxes is the same, 86\% across all categories.

Fig~\ref{fig:masks} shows qualitative comparisons of annotations from the proposed method to the ground truth labels for each category. For larger objects, such as the boxes and concrete barriers with less fine details, both the bounding boxes and the segmentations are qualitatively very close to the ground truth, as can also be seen by the quantitative results in Table~\ref{tab:iou}. For the more challenging categories with smaller objects or many fine details, the produced segmentations and bounding boxes are not perfect. For the food item category, both the segmentations and bounding boxes have considerable variance, suggesting that the reconstruction of the object meshes does not always succeed. For the scooter category, there is considerably more variance in the bounding box error compared to the variance in segmentation, suggesting that small errors in the segmentation masks cause a larger error in the bounding boxes. Some of the categories (bottle, pole) suffer categorically from the reconstruction not being able to reconstruct the top part of the object in the mesh, but it does not cause a corresponding issue in the bounding boxes as with the scooter category.

The 3D bounding box accuracy is hard to evaluate, as we can't obtain ground truth information on the object position and orientation. See the accompanying video for videos of the bounding boxes rendered on some video sequences. A selection of bounding boxes are shown in Figure \ref{fig:boxes}.

\begin{figure*}[t]
    \includegraphics[width=\linewidth]{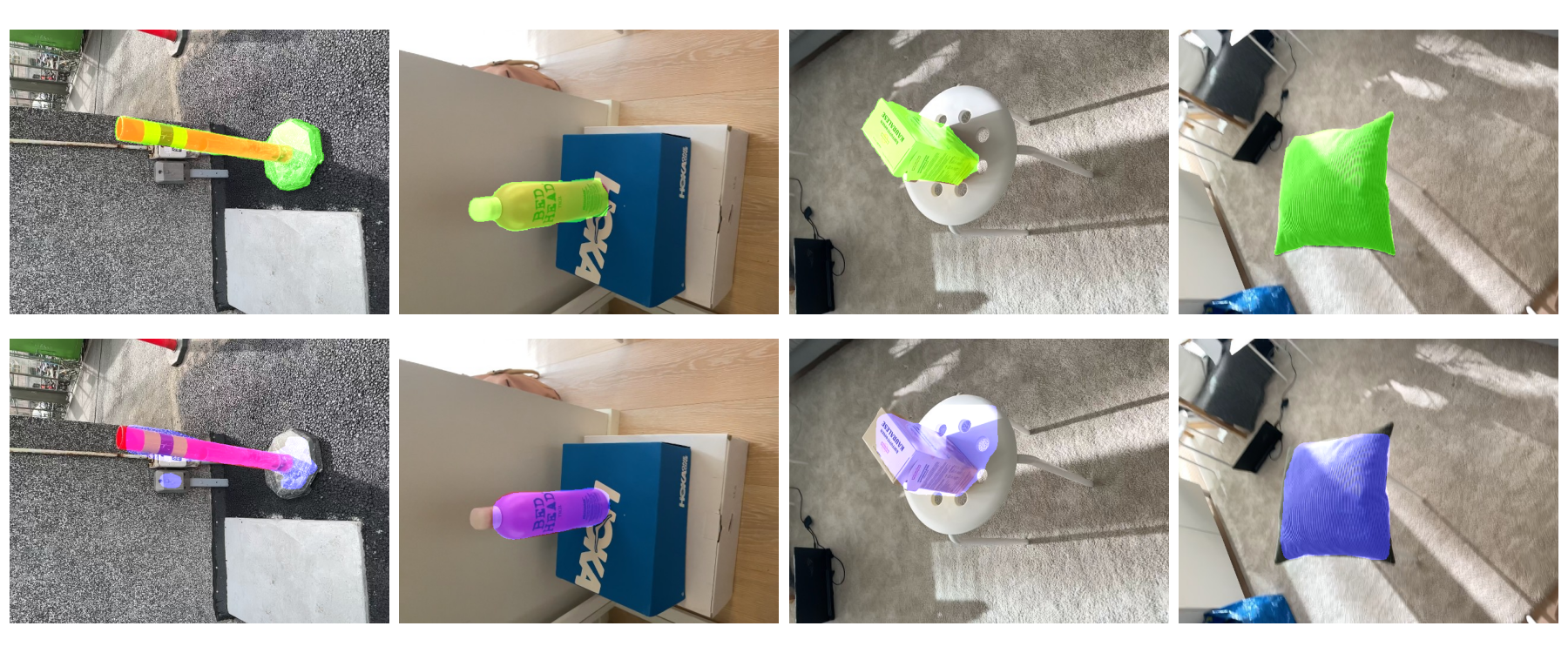}
    \centering
    \caption{Qualitative examples of model predictions (bottom row) compared to ground truth labels (top row)}
    \label{fig:pred-masks}
\end{figure*}

\subsection{Labeling efficiency}

Using our 3D annotation tool, it took us 16 minutes and 45 seconds to label all of our 34 scenes containing 47 objects. Giving an average of 21 seconds per object. If we assume one object per scene and an average of 1665 images per scene in our dataset, we get a labeling throughput of 3442 images per minute.

For the ground truth dataset, we measured the time it took us to label images using the traditional method where images are annotated one by one. We measured a session of labeling 200 images randomly picked from our 34 scenes. It took us 2 hours and 43 minutes to label those 200 images. Averaging at 1 minute and 7 seconds per image. Although, it should be noted that there is significant deviation from image to image depending on the amount of detail on an object, the complexity of the contour from the specific viewpoint and the amount of objects there are in the scene. Labeling a scooter is much more time consuming than labeling a cardboard box.

It is thus evident, that a 3D labeling approach is much more efficient. However, it does come at a slight cost in label quality. Also, raw throughput is not an entirely fair comparison, as the images coming from the 3D scene annotation pipeline do correlate along the temporal dimensions. If you labeled individual images, you could pick less correlated examples that are further apart and only label those, but these might take longer to capture. However, the benefits of labeling all dimensions still stand.

\subsection{Downstream task}

As we showed, the produced per-pixel and 2D bounding box annotations are not quite perfect for all categories compared to ground truth labels. To evaluate the effect of the label quality on a downstream task, we trained a state-of-the-art instance segmentation model \cite{wu2019detectron2} using the full dataset, with 8 categories. After training, we collected and manually annotated an additional 100 images for running predictions and computing the IoU score to evaluate performance. The mean and standard deviation are shown in Table~\ref{tab:iou-train} for each category and qualitative examples are shown in Fig~\ref{fig:pred-masks}. Overall the performance is slightly below the original labels, with a mean IoU of 0.81 for both the per-pixel segmentation and bounding boxes. We notice that for the larger objects (barrier, box, pillow), the results are comparable or better than the label metrics in Table~\ref{tab:iou}, suggesting that in some cases the errors in labels average out in model training. The only category that performs significantly worse in model inference compared to the labels is the scooter category, where both the segmentation (0.81 vs 0.63) and bounding box (0.84 vs 0.74) errors are larger. This again highlights the fact that if small details are consistently missing from the labels, it is challenging for the model to recognize them as a part of the object, especially if the unseen examples are very different from what the model was trained with. Overall, the result suggests that training well performing instance segmentation models does not require perfect labels, but more data and higher fidelity labels are needed in cases where objects are small or have many distinct details.

\section{CONCLUSIONS}

In this paper, we showed that using SLAM and a 3D reconstruction pipeline is a valid approach for creating high quality 2D and 3D annotations of arbitrary objects for both indoor and outdoor settings. The presented 3D annotation method is able to produce labels several orders of magnitude faster than a user labeling semantic segmentation masks the traditional way. While these semi-automatically created labels are slightly less accurate than their artisanal counterparts, we showed that the reduced accuracy can be tolerated on downstream tasks.

As our segmentation masks are not perfect, future work could very well focus on how to combine the information available from the reconstruction pipeline, color and depth readings to come up with a slightly improved segmentation mask. Annotation could also further be sped up by automatically detecting objects in scenes.

Other avenues for future work include adding a time dimension to the tool, making the scene based annotation approach work on dynamic scenes with moving objects. Once the objects have been annotated in a single depth frame, features within the box could be computed and associated into tracks across all the frames in the sequence. These tracks could then be used to create a voxel grid with a time dimension to capture the evolution of the scene over time. Furthermore, the current method only uses inputs from one camera. Several simultaneous viewpoints could be used to improve the reconstruction performance, annotation throughput, and capture more information about the scene.

\addtolength{\textheight}{-12cm}   

\bibliographystyle{IEEEtran}
\bibliography{references}

\end{document}